\pdfoutput=1

\documentclass[11pt]{article}

\usepackage{acl}

\usepackage{times}
\usepackage{latexsym}

\usepackage[T1]{fontenc}

\usepackage[utf8]{inputenc}

\usepackage{microtype}

\usepackage{inconsolata}

\usepackage{times}
\usepackage{latexsym}
\usepackage{graphicx}
\usepackage{tikz}
\usepackage{booktabs}
\usepackage[breakable]{tcolorbox}
\usepackage{multirow}

\newcommand{\Asquare}{%
    \begin{tikzpicture}
        \fill[red] (0,0) rectangle (0.2cm,0.2cm);
    \end{tikzpicture}%
}

\newcommand{\Bsquare}{%
    \begin{tikzpicture}
        \fill[green] (0,0) rectangle (0.2cm,0.2cm);
    \end{tikzpicture}%
}

\newcommand{\Csquare}{%
    \begin{tikzpicture}
        \fill[blue] (0,0) rectangle (0.2cm,0.2cm);
    \end{tikzpicture}%
}

\title{StyleChat: Learning Recitation-Augmented Memory in LLMs \\for Stylized Dialogue Generation}

\author{Jinpeng Li$^{1}$\thanks{\ \ Equal contribution.}, Zekai Zhang$^{1}$\footnotemark[1], Quan Tu$^2$, Xin Cheng$^1$, Dongyan Zhao$^{1,3}$\thanks{\ \ Corresponding authors: Dongyan Zhao and Rui Yan.}, Rui Yan$^{2}$\footnotemark[2] \\
$^1$Wangxuan Institute of Computer Technology, Peking University\\
$^2$Gaoling School of Artifical Intelligence, Renmin University of China\\
$^3$State Key Laboratory of Media Convergence Production Technology and Systems\\
  \texttt{lijp.pku@gmail.com,justinzzk@stu.pku.edu.cn}\\
  \texttt{zhaody@pku.edu.cn, ruiyan@ruc.edu.cn} \\
}

\begin{document}
\maketitle
\begin{abstract} 
Large Language Models (LLMs) demonstrate superior performance in generative scenarios and have attracted widespread attention. 
Among them, stylized dialogue generation is essential in the context of LLMs for building intelligent and engaging dialogue agent.
However the ability of LLMs is data-driven and limited by data bias, leading to poor performance on specific tasks. In particular, stylized dialogue generation suffers from a severe lack of supervised data.
Furthermore, although many prompt-based methods have been proposed to accomplish specific tasks, their performance in complex real-world scenarios involving a wide variety of dialog styles further enhancement.
In this work, we first introduce a stylized dialogue dataset \textbf{StyleEval} with 38 styles by leveraging the generative power of LLMs comprehensively, which has been carefully constructed with rigorous human-led quality control.
Based on this, we propose the stylized dialogue framework \textbf{StyleChat} via recitation-augmented memory strategy and multi-task style learning strategy to promote generalization ability.
To evaluate the effectiveness of our approach, we created a test benchmark that included both a generation task and a choice task to comprehensively evaluate trained models and assess whether styles and preferences are remembered and understood.
Experimental results show that our proposed framework StyleChat outperforms all the baselines and helps to break the style boundary of LLMs.
\end{abstract}

\section{Introduction}

\begin{figure}[t] 
\centering 
\includegraphics[width = 0.45\textwidth]{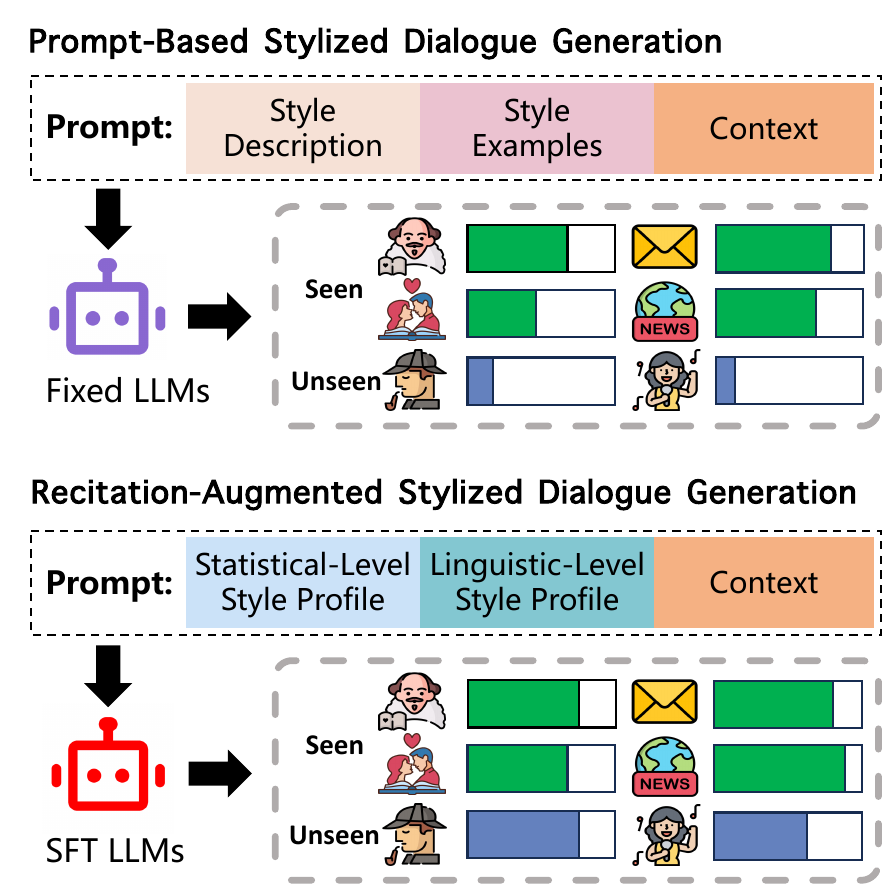}
\caption{Examples of stylized dialogue generation by different LLMs. The progress bar represents the quality in the particular style.}
\label{Fig:example}
\end{figure}

Large language models (LLMs) have made considerable advancements in the field of natural language processing~\cite{openai,brown2020language,zhang2023survey}. These models demonstrate a deep comprehension of the context and semantics of complex instructions and are capable of generating coherent and logical texts \cite{ouyang2022training}. Especially in the field of dialogue, the integration of large language models has enabled machines to interact with humans in a more natural, personalized, and stylized manner \cite{lv2023envisioning,lv2023dialogps,tu2023characterchat}, heralding a new phase in the evolution of artificial intelligence.

Stylized dialogue generation is crucial in the development of intelligent and engaging dialogue agents \cite{zheng2021stylized,li2023stylized} in the era of LLMs. Nonetheless, this task is challenged by the limited availability of supervised data correlating contexts and responses with the desired styles \cite{Gao2019StructuringLS,li2023stylized}. 
In particular, it is more difficult to collect parallel corpus for abstract, multilayered or dynamically derived styles. 
Existing works have typically relied on pseudo data constructed using back translation, thereby resulting in low-quality data that fail to account for the variability and complexity of individual language styles \cite{Su2020DiversifyingDG,Zheng2020StylizedDR,Li2021StylizedDG,li2023stylized}. The result is that the model generated dialogues tends to be overly standardized, lacking individuality and diversity.
Additionally, despite the introduction of prompt-based LLMs specifically designed for certain tasks, their performance in complex real-world scenarios necessitates improvement \cite{zeng2023agenttuning}. Particularly, when encountering domain data or new style not seen during the pre-training phase, the generalization ability of LLMs dealing with complex instruction significantly declines.

To address these challenges, we exploit the generative capacity of LLMs, combined with statistical and linguistic perspectives, to construct a large-scale dataset, named \textbf{StyleEval}. 
This dataset consists of stylized dialogues with style profiles, contributing to the creation of custom dialogue agent. To our knowledge, this is the first large-scale dataset for stylized dialogue generation, incorporating 38 styles and 24,728 dialogues. 
Our process begins with the collection of well-known styles from various genres, utilizing GPT-4 to generate statistical-level style profile that includes descriptions and examples. We then extract linguistic-level style profile from these examples based on linguistic knowledge. After initial pre-processing, we invite annotators to evaluate the quality of the dialogues.
Furthermore, we aim to enhance the style generalization ability of the LLM without compromising its overall functionality. 
However, direct prompting methods of LLMs face challenges to generalize to new styles, as depicted in Figure \ref{Fig:example}. 
We adapt the model to generate responses for styles it has not previously encountered. 
To address this, we propose the \textbf{StyleChat} framework, which introduces a style thought chain, enabling models to generate style profiles before responding via a recitation-augmented memory strategy. This memory consists of two stages: recite then respond during training, and recall then respond during inference. This approach also encourages StyleChat to learn how to derive unseen style profiles, thereby improving generalization. Besides, we further enhance the style derivation ability by implementing multi-task style learning to increase activation of style abilities through a style transfer dataset.

Comprehensive experiments conducted on StyleEval demonstrate that our approach considerably enhances the performance of LLMs in stylized dialogue generation. 
StyleChat accurately captures the essence of various styles and generates dialogue content that is rich in stylistic elements. By utilizing meticulously constructed supervised data and the recitation-augmented memory strategy, we can effectively transcend the limitations of LLMs on specific tasks, thereby improving their performance in novel styles.
We also discuss the advantages of our approach extensively in Appendix. In summary, our contributions can be summarized as follows:
\begin{itemize}
\item We construct a large-scale, high-quality dataset, StyleEval, for the stylized dialogue generation. This dataset comprises 24,728 parallel stylized dialogue turns covering 38 diverse styles, serving as a crucial prerequisite for successful style-playing.
\item We introduce a recitation-augmented memory strategy for stylized dialogue generation, which motivates StyleChat to learn to derive unseen style profiles for better generalization.
\item We conduct extensive experiments on various large language models under both in-domain and out-of-domain settings using StyleEval, demonstrate that our proposed framework, StyleChat, outperforms all baseline models.
\end{itemize}

\section{Related Work}
\subsection{Stylized Dialogue Generation}
Stylized dialogue generation represents a significant research direction within the field of intelligent dialogue systems, focusing on generating dialogue imbued with specific stylistic characteristics. Initial approaches primarily depended on the utilization of latent variables within the hidden state space~\cite{gao2019structuring}
Some researchers have attempted to integrate pseudo data into existing corpora via back translation~\cite{sennrich2015improving,he2016dual,Zheng2020StylizedDR,Li2021StylizedDG}. 
With the advent of pre-trained models, StyleDGPT~\cite{yang2020styledgpt} employs both a style language model and a style classifier to provide style signals. Nonetheless, these methods often struggle to capture and reproduce complex linguistic style features, leading to generated dialogues that may be overly mechanical and homogeneous. Furthermore, these methods demonstrate limited generalization capabilities when adapting to new or unseen styles. Thus, our research seeks to surmount these limitations inherent in traditional stylized dialogue methodologies, aiming to enhance the performance and quality of stylized dialogue through the construction of supervised data and fine-tuning, leveraging the instruction comprehension and generation capabilities of large language models.

\subsection{Domain-Specific LLMs}
To augment the performance of models in specific domains (e.g., medical, legal, character-based, etc.), supervised fine-tuning (SFT) and in-context learning (ICL) have emerged as dominant methodologies\cite{raffel2020exploring,dong2022survey}. 
Domain-specific large language models have been extensively explored by researchers~\cite{singhal2022large,cui2023chatlaw,tu2023characterchat}. 
These models are tailored to understand and generate content within a specific domain, enabling them to capture domain-specific nuances and terminology. 
In-context learning capitalizes on the inherent ability of large language models to swiftly adapt to a specific context or task with a few examples~\cite{garg2022can}. By supplying contextually relevant instructions (e.g., prompts or dialogues), this approach partly facilitates the generalization and effective performance within a specific domain~\cite{radford2018improving,Begu2023LargeLM}. Nonetheless, its performance can be constrained by the number and quality of examples, and it may lack the accuracy required for complex or fine-grained style transformations. While these studies concentrate on adapting models to various domains, existing domain-specific LLMs may not fully meet the requirements due to the diversity and complexity of stylized dialogue, indicating the need for further research and optimization.

\section{Methodology}
\subsection{Dataset Construction}
\textit{Design Principles}: LLMs are pre-trained on an extensive range of texts, including various style corpora. This expansive training embeds a rich repository of stylistic knowledge within parameters, thereby offering significant potential for stylized dialogue generation. However, exploiting the style potential of LLMs often necessitates accurate style definitions and specific strategies. Therefore, we focuses on two primary objectives:
\textbf{1) Efficient alignment of LLMs to a certain style.} To effectively tailor LLMs to a particular style, we employ style definitions from both statistical and linguistic perspectives, as illustrated in Figure~\ref{Fig:dataset}. 
\textbf{2) Activate their style-related abilities for better generalization.} To optimally activate the style-specific capabilities of LLMs, we strategically design the data distribution of our two style-centric tasks, stylized dialogue generation and text style transfer. 

\subsubsection{Statistical-Level Style Profile}\label{sec:Statistical-Level Style Profile}
As outlined in \cite{Jin2020DeepLF}, conventional deep learning approaches typically define style from a statistical or data-driven perspective.
These methods involve training models on large corpora of texts or responses with very different styles, allowing models to learn characteristics of the various styles independently.
Consistent with these approaches, we employ GPT-4 as a style agent to create a statistical-level style profile. Specifically, we task the agent to generate a comprehensive description of a specific style, leveraging its extensive, statistically-informed understanding of styles. Subsequently, agent is used to generate a series of sentences representative of the specified style. To ensure the relevance and accuracy of these examples, we implement a post-selection phase. During this stage, human annotators meticulously select sentences that most effectively embody the core characteristics. Given the robust in-context learning abilities of LLMs, we limit the number of example sentences to four, aiming for a precise yet efficient statistical definition of style.

\subsubsection{Linguistic-Level Style Profile}
\label{sec:Linguistic-Level Style Profile}

In addition to the statistical perspective, we also delve into the linguistic perspective of style. 
We argue that representing style through a large collection of sentences can be inadequate. 
It lacks explicit guidance on how to produce stylized sentences and is resource-intensive, especially when it comes to generalization, as gathering extensive corpora for new styles is costly. 
Different from conventional style language models, LLMs demonstrate good linguistic understanding ability, as highlighted by \cite{Begu2023LargeLM}. 
Therefore, we propose a re-evaluation of the concept of style from a linguistic perspective and its integration with the statistical level style profile.
This combined approach is designed for more efficient and accurate style definition and enhanced generalization capabilities. Specifically, we adopt the definition provided by ~\cite{Kumar2022StyleAS}, \textit{The style has been analysed in such terms as rhetorical situation and aim, diction or choice of words,  type of sentence structure and syntax,and  the density and kinds of figurative languages.} 
Based on this, we decompose style into the following four attributes for more precise and comprehensive guidance:

\begin{figure}[t]
\centering
\includegraphics[width = 1 \columnwidth]{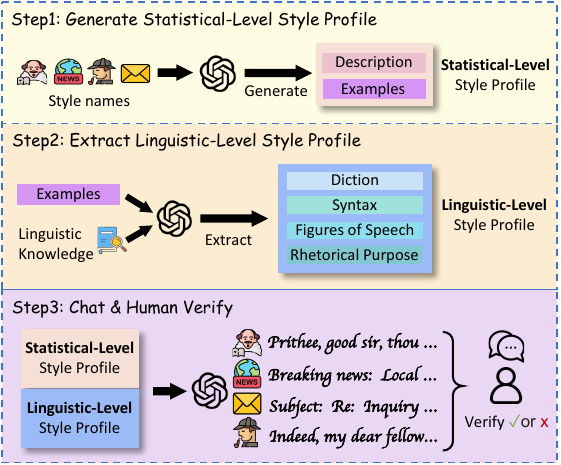}
\caption{The workflow for developing StyleEval. LLM is  employed to generate statistical-level style profile for a certain style. Then to extract linguistic-level style profile based on examples and linguistic knowledge. Finally, we produce stylized dialogue with context and style profile, verified by human to guarantee quality.}
\label{Fig:dataset}
\end{figure}

\begin{figure*}[t]
\centering 
\includegraphics[width = 0.95 \textwidth]{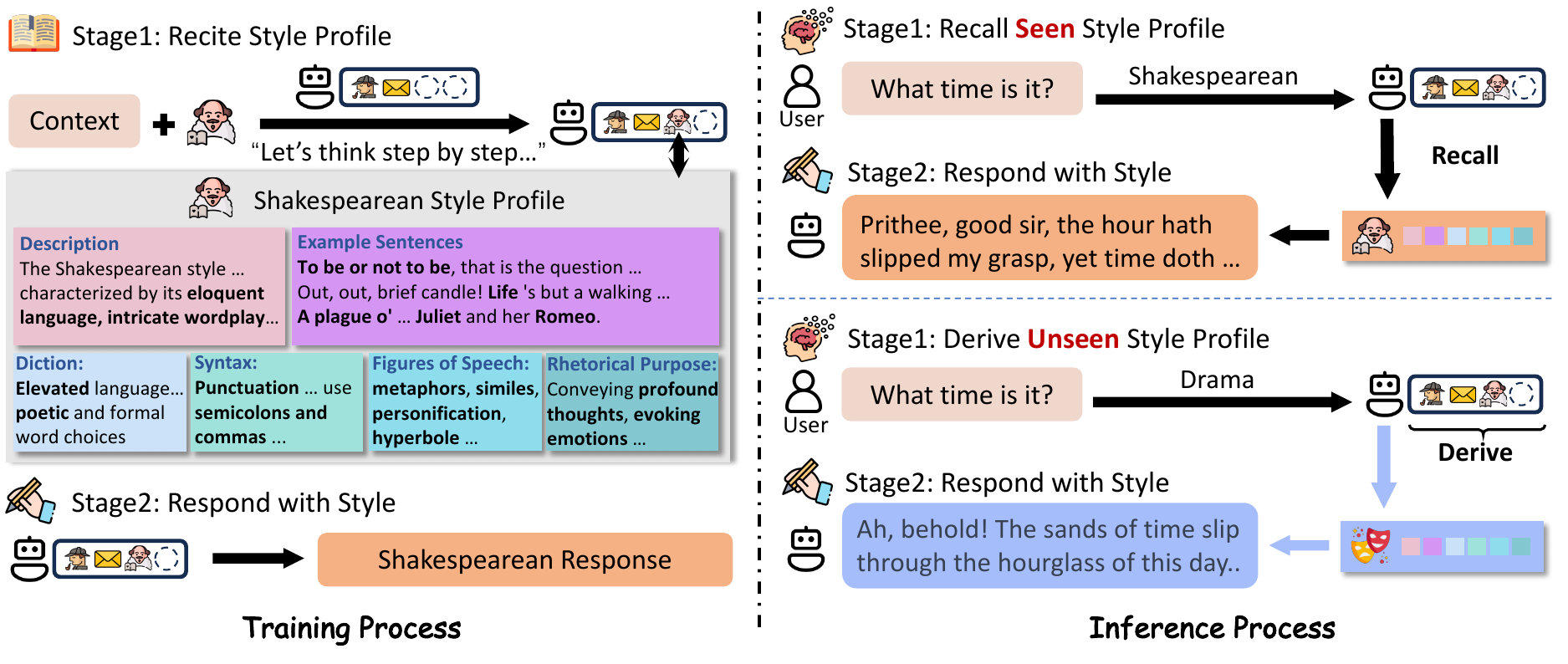}
\caption{The overview of our proposed StyleChat Framework. During training phase, our model is instructed to first recite the style profile then respond with reference to the recited style profile. During Inference, model recalls or derives profiles from parametric memory and then respond with style. Our two-stage framework teaches model to learn implicit Chain of Thought process, resulting in better generalization abilities through chains of style thoughts.}
\label{Fig:framework}
\end{figure*}

\noindent \textbf{\texttt{Diction}} is the choice of words and style of expressions, are the most basic elements of style~\cite{Kumar2022StyleAS,Jin2020DeepLF}. 
For example, the use of complex and technical vocabulary is apt for academic papers (arXiv styles), but might be less appealing in a novel targeted at a general audience.

\noindent \textbf{\texttt{Syntax}} is the arrangement of words and phrases to create well-formed sentences in a language \cite{Kumar2022StyleAS}. 
``To be or not to be, that is the problem" in Shakespeare is a classic illustration of syntax influence.
Altering this to a more standard syntax, such as ``The question is whether to be or not to be", diminishes its Shakespearean essence.

\noindent \textbf{\texttt{Figures of Speech}} is the creative uses of language where words take on a non-literal meaning as described by \cite{Konig2016AGO}. 
These include devices like metaphors, similes, personification, and hyperbole. 
For instance, in the Poems style, William Wordsworth's line ``I Wandered Lonely as a Cloud" employs a simile to liken the narrator's solitude to a cloud, conjuring a feeling of freedom.

\noindent \textbf{\texttt{Rhetorical Purposes}} refer to the objectives a speaker aims to accomplish through communication \cite{McDonald1985ACT}. In Questionnaire style, the rhetorical purpose is information gathering, thus necessitating many interrogative sentences. Conversely, the goal is to entertain in Humor style, requiring humorous content.

The comprehensive process for creating a style profile is illustrated in Figure \ref{Fig:dataset}.
Initially, GPT-4 serve as style agent and is prompted to generate a concise overall description and four representative examples for the given style, forming the statistical-level style profile. 
We then integrate linguistic knowledge to guide agent in extracting relevant linguistic attributes from these examples, leading to the development of the linguistic-level style profile. It is worth highlighting the distinction between ``style" and ``persona" or ``character" here. While the latter emphasizes content or experiences, we focus on stylistic content along with linguistic attributes. Furthermore, certain styles (e.g., Email, Lyrics, News) should be classified as a style rather than a persona or character.

\subsubsection{Multi-Task Datasets for Style Activation}
\label{sec:Multi-Task Datasets for Style Activation}
In this section, we outline the development of StyleEval, which encompasses two style-centric tasks: Stylized Dialogue Generation and Text Style Transfer. 
For the stylized dialogue generation, we build upon multi-level style profiles discussed in the previous sections. 
By engaging GPT-4 with style profiles corresponding to certain styles and dialogue, we construct pairs of contexts and stylistic responses. 
These pairs subsequently serve as training data for our model in a multi-turn dialogue setting. Guided by insights from \cite{Chan2022DataDP}, which suggest that datasets featuring a combination of several principal clusters along with a multitude of rare instances enhance the model's generalization capabilities, we structure our dataset accordingly.
We curate a collection featuring 3,532 examples for 4 main styles, supplemented by 400 examples for 23 less prevalent styles, aiming to optimize the model’s generalization potential. 
For the text style transfer, we utilize GPT-4 to obtain transfer instances between any pair of styles in four primary styles, totalling 600 pairs of data. 
Despite limited amount of data, we demonstrate the efficacy of multi-task learning in enhancing generalization and further in the context of previously unseen styles in Section \ref{exp:performance} . 
We collate the aforementioned prompts in the Appendix.

\subsection{Domain-specific Alignment Strategies}

To optimize large language models for specific styles and augment their adaptability across various styles, we introduce StyleChat as shown in Figure \ref{Fig:framework}, which incorporates recitation-augmented memory and multi-task style learning strategies. 
We explain our motivation using the process of an apprentice cook learning to prepare diverse cuisines.
Imagine an expert chef is instructing an apprentice on how to create dishes with unique flavors (styles). 
The initial task of apprentice entails \textit{reciting} recipes (style profiles) for a specific flavor. 
During the cooking process, the apprentice \textit{recalls} these recipes and meticulously follows their instructions. 
To further the apprentice's comprehension of different culinary styles, the chef challenges him to apply \textit{varied flavors} (style transfer) to \textit{same ingredients}. 
This method enables the apprentice to learn differences between flavors, thus grasping the essence of recipes, and become adept at adapting to new recipes. Correspondingly, we propose the recitation-augmented memory strategy coupled with a multi-task style learning process. We will introduce these concepts in the following subsection.

\subsubsection{Recitation-Augmented Memory} 
\label{sec:Recitation-Augmented Memory}
Departing from the conventional approach that relies on prompts, we propose a recitation-augmented memory strategy to enhance the style capabilities of LLMs for better generalization. 
Inspired by the concept of Chain of Thought (CoT) \cite{Wei2022ChainOT} prompting, we structure our stylized dialogue generation process as a two-stage framework. 
Specifically, during the training phase, our model is first instructed to recite the relevant style profile triggered by the prompt ``Let's think step by step" followed by additional guiding prompts. 
Subsequent to this recitation, the model is tasked with generating a response that is consistent in style with the recited profile. 
To formulate, given a style $\mathcal{S}$ with its corresponding style profile denoted as $\mathcal{P}$, the dialogue context is defined as $\mathcal{C}=\{x_1,y_1,x_2,y_2...x_k\}$, where $x_k$ represents the $k$-th utterance from the first person, and $y_k$ represents the response from the second person in style $\mathcal{S}$, it can be expressed as:
$$
p(y_k|\mathcal{C},\mathcal{S}) = \sum p(y_k|\mathcal{C},\mathcal{P}) \cdot p(\mathcal{P}|\mathcal{C},\mathcal{S}),
$$
and the dialogue generation loss $\mathcal{L}_{SD}$ can be formulated as:
$$
\mathcal{L}_{SD} = \log p(y_k|\mathcal{C},\mathcal{S}).
$$
During the training phase, our model is compelled to first recite the correct style profile, followed by outputting the stylized response. 
This is achieved by appending the style profile to the appropriate stylized response as part of the label. 
In the inference phase, we adopt two different settings, seen styles and snseen styles.
For seen styles, StyleChat first recalls the given style profile and then responds a response in reference to the profile. Specifically, we employ the input and output separate token $\texttt{SEP}$, placing the style profile after $\texttt{SEP}$. 
This is equivalent to the model being forced to output the correct style profile first before generating the stylized responses. 
For unseen styles, StyleChat utilize its recitation-augmented memory capability to derive the style profile and then generate a stylized response based on it.
This instructs LLMs to model style-related tasks via an implicit Chain of Thought process. 
In addition to the traditional training with stylized dialogue as a label, by reciting style profiles from memory, LLMs gained a deeper understanding of the styles in the parameter space, thus improving their ability to generalise across styles.

\subsubsection{Multi-Task Style Learning}\label{sec:MultiTask Style Learning}

To enhance the comprehensive style understanding of our model and thereby improve stylized dialogue generation tasks, we propose a multi-task style learning framework.
This approach involves training the model not only on the stylized dialogue generation as stated above but also on the text style transfer.
We demonstrate the effectiveness of multi-task style learning in Section \ref{exp:performance}. 
For the style transfer, suppose we want to transfer a sentence $t$ in style $\mathcal{S}_{1}$ to $t^{'}$ in $\mathcal{S}_2$. We formulate the loss as follows:
$$
\mathcal{L}_{ST} = \log p(t^{'}|t,\mathcal{S}_1,\mathcal{S}_2),
$$
thus, the overall supervised loss $\mathcal{L}_{SFT}$ can be expressed by:
$$
\mathcal{L}_{SFT} = \lambda_{SD} \cdot \mathcal{L}_{SD} + \lambda_{ST} \cdot \mathcal{L}_{ST},
$$
where $\lambda_{SD}$ and $\lambda_{ST}$ denote the corresponding loss weight for stylized dialogue and style transfer.

\section{Experiments} 
\subsection{Experimental Setup} 
\label{exp:setup}

\textbf{Implementation  Details.}
We implemented StyleChat with the basis of LLaMA2-7B-chat~\footnote{https://huggingface.co/meta-llama/llama-2-7b-chat-hf} using Low-Rank Adaptation (LoRA)~\cite{hu2021lora}. 
All experiments are conducted on a single A800 GPU. For LoRA, we set the rank $r$ to 256 and alpha $\alpha$ to 128, training across eight layers of all projection parameters, rendering 8.7\% of the total parameters trainable. The coefficients $\lambda_{SD}$ and $\lambda_{ST}$ are all set to 1.0.
StyleChat is trained for six epochs with an initial learning rate of $5e-5$ and a cosine learning rate scheduler. 
We utilize the batch size of 32 and train for one day. 

\textbf{Dataset Statistic.} 
\label{exp:statistic}
The StyleEval dataset is meticulously divided into distinct training and test sets, as shown in Table~\ref{Tab:statistic}, and the details of each style are presented in the Appendix. 
The selected styles encompass a broad spectrum of communication styles, providing the model with a rich variety of learning samples to enhance its accuracy in style imitation and generation. 
We sample dialogues from DailyDialog Dataset~\cite{Li2017DailyDialogAM} and provide GPT-4 with contexts, style profiles to generate stylized response as labels. 
Specifically, the training set contains 23,328 stylized dialogue instances across 27 distinct styles. The test set encapsulates 1,000 stylized dialogue spread across 38 diverse styles. 
Notably, the test set includes 11 styles that are not encountered during training, purposefully included to evaluate the generalization capabilities in an out-of-domain setting. 
Inevitably, the randomness of LLMs generation can impact data quality. To mitigate this, we invite human annotators to assess the coherence and quality of the conversations and to eliminate any problematic instances.

\begin{table}[t]
\centering
\resizebox{1 \columnwidth}{!}{
\begin{tabular}{lcccc}
\toprule
\multirow{2}{*}{} &  \multicolumn{2}{c}{\textbf{Training}} & \multicolumn{2}{c}{\textbf{Test}} \\
\cmidrule{2-5}
& Dialogue & Transfer &  Generation  & Choice \\
\midrule
\# Instances & 23,328 & 600 & 1,000 & 400 \\
\midrule
\# Styles & 27 &  4  &  38 & 38 \\
\midrule
Avg. Tokens & 23.0 & 32.3 & 32.2 &  51.2 \\
\midrule
Avg. Turns & 3.25 & 1.00 & 2.99 & 2.82 \\
\midrule
Avg. Profiles & 1.00 & -  & 0.54 & -   \\
\bottomrule
\end{tabular}}
\caption{The statistics of StyleEval dataset.}
\label{Tab:statistic}%
\end{table}%

\begin{table*}[t]
\centering
\resizebox{2 \columnwidth}{!}{
\begin{tabular}{lcccccccccc}
\toprule
Method & BLEU-1 & BLEU-2 & BLEU-3 & BLEU-4 & Rouge-1 & Rouge-2 & Rouge-L & Distinct-1 & Distinct-2 & Length \\
\midrule
LLaMA2-7B-Chat & 24.76 & 4.53 & 1.52 & 0.77 & 18.72 & 3.23 & 16.61 & 16.05 & 54.89 & 77.30 \\
LLaMA2-13B-Chat & 24.60 & 4.85 & 1.65 & 0.83 & 20.28 & 3.86 & 17.87 & 15.66 & 56.82 & 89.99 \\
ChatGPT & 32.90 & 8.46 & 3.64 & 1.96 & \textbf{25.30} & \textbf{6.39} & \textbf{22.76} & 14.74 & 56.38 & 67.44 \\
\midrule
StyleChat(7B) & \textbf{42.03} & \textbf{12.09} & 5.49 & \textbf{3.09} & 21.63 & 5.71 & 19.23 & \textbf{22.29} & \textbf{65.91} & 32.43 \\
\quad w/o Transfer & 41.46 & 12.08 & \textbf{5.51} & 3.09 & 21.49 & 5.63 & 19.07 & 22.18 & 65.74 & 32.91 \\
\quad w/o Profile & 41.40 & 11.78 & 5.36 & 3.04 & 22.35 & 5.82 & 19.81 & 21.54 & 64.77 & 34.30 \\
\quad w/o Recite & 41.42 & 11.85 & 5.33 & 2.99 & 22.57 & 6.01 & 19.99 & 21.36 & 64.93 & 35.16 \\
\bottomrule
\end{tabular}}
\caption{Automatic evaluation results of StyleChat, baselines and ablation models on test dataset of StyleEval. w/o Transfer means we only use the stylized dialogue data for training while discarding style transfer task. w/o Profile means we do not provide style profile for LLMs. w/o Recite means we do not use our recitation-augmented memory and append style profile in prompt.}
\label{Tab:main}
\end{table*}

\textbf{Compared Baselines.} \label{exp:baselines}
To verify the effectiveness of our proposed method, we conduct a comprehensive comparison of baseline models, including a spectrum of large language models distinguished by parameter sizes and types.
The models included in our evaluation are LLaMA2-7B-chat (LLaMA2-7B.), LLaMA2-13B-chat (LLaMA2-13B.)~\cite{Touvron2023Llama2O}, ChatGPT \cite{openai}, Baichuan2-7B \cite{Yang2023Baichuan2O}, ChatGLM3-6B \cite{Du2021GLMGL} and Vicuna-7B-v1.5 \cite{chiang2023vicuna}. 
Through this comparative analysis, we aim to highlight the relative strengths and weaknesses of each model, providing insights into how different aspects of a model influence generation, especially within the context of stylized dialogue.

\textbf{Evaluation Metrics.} \label{exp:metrics}
In alignment with previous studies, we employ reference-based evaluation metrics ROUGE~\cite{lin2004rouge} and BLEU~\cite{Papineni2002BleuAM} to measure the n-gram overlap between generated and reference responses for automatic evaluation. 
The Distinct~\cite{Li2015ADO} is used to measure the proportion of unique n-grams in the generated responses. 
In addition to automatic metrics, we conduct both LLM and human evaluations to assess the quality of generated stylized responses based on Relevance, Coherence, and Style: 
1) Relevance measures how well the response aligns with the given context. 
2) Coherence measures the extent to which the context and response form a coherent body of information.
3) Style measures the degree to which the response reflects the desired style. 
For the LLM evaluation, we use GPT-4 as judger to rate the responses, providing it with the responses, contexts, and specific criteria for each dimension.
In the case of human evaluations, we instruct our annotators to use the same criteria, as presented in the Appendix.

\begin{table}[t]
\centering
\small
\begin{tabular}{lccc}
\toprule
Method & Relevance & Coherence & Style \\
\midrule
LLaMA2-7B-Chat & 2.89 & 3.15 & 3.91 \\
LLaMA2-13B-Chat & 3.63 & 3.86 & 4.27 \\
ChatGPT & 4.49 & 4.58 & 4.47 \\
\midrule
StyleChat & 4.68 & \textbf{4.81} & \textbf{4.69} \\
\quad w/o Transfer & 4.67 & 4.69 & 4.44 \\
\quad w/o Profile & \textbf{4.75} & 4.77 & 4.50 \\
\quad w/o Recite & 4.72 & 4.75 & 4.48 \\
\bottomrule
\end{tabular}
\caption{GPT-4 evaluation results on test dataset. We employ GPT-4 with detailed rating criterias as judger to rate generated stylized responses in terms of Relevance, Coherence and Style.}
\label{Tab:gpt-4}
\end{table}

\begin{figure}[t]
\centering
\includegraphics[width = 1\columnwidth]{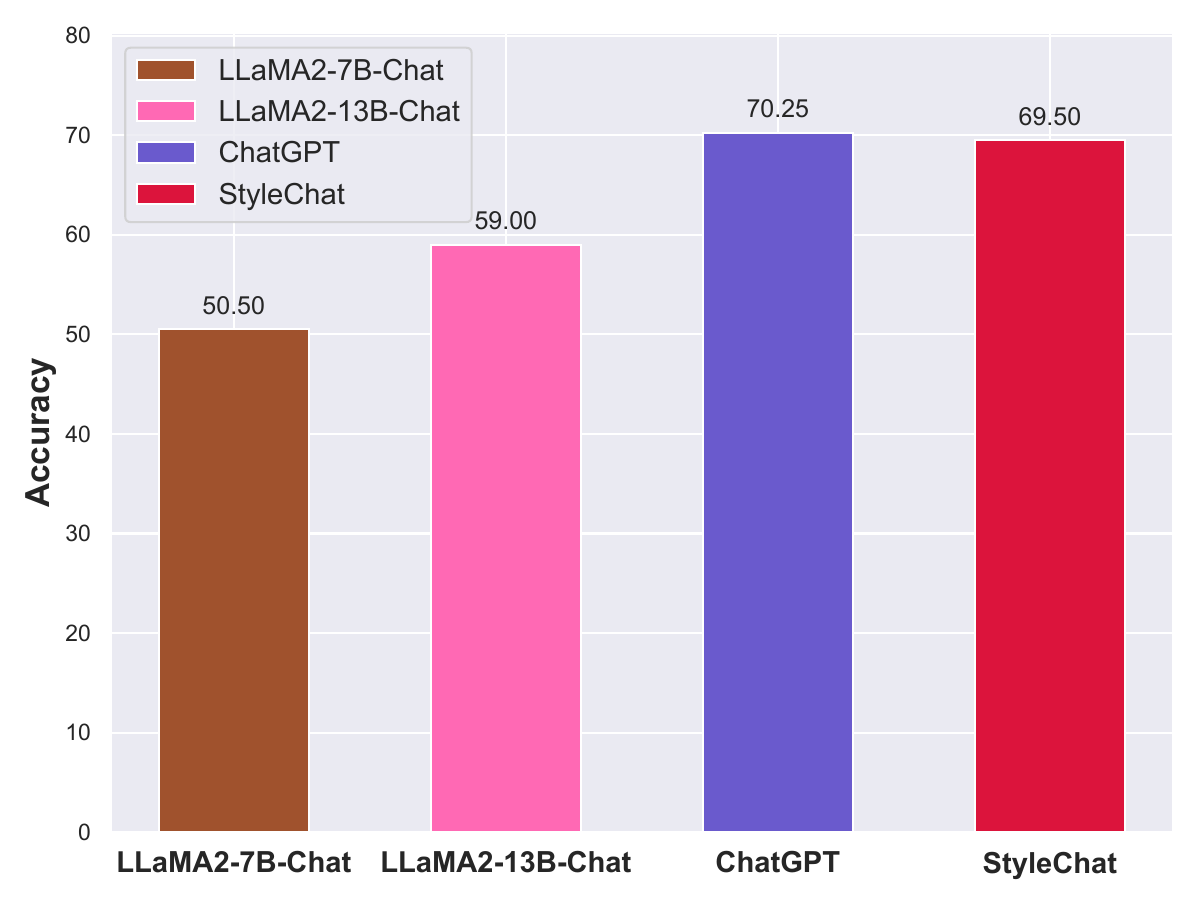}
\caption{The multiple choice evaluation, where y-axis represents accuracy and x-axis lists different models.}
\label{Fig:mainbar_choice}
\end{figure}

\subsection{Results and Analysis} 
\textbf{Overall Performance.} 
\label{exp:performance}
The automatic results and the GPT-4 evaluation results are summarized in the Table~\ref{Tab:main} and Table~\ref{Tab:gpt-4}, respectively.
Overall, our method achieves the highest BLEU and Distinct scores on the StyleEval dataset, which shows the superiority of our approach.
This significant achievement validates the effectiveness of our method in capturing and reproducing stylized nuances within dialogue generation, demonstrating the model's proficiency in producing text that aligns closely with the desired stylistic characteristics through the recitation-augmented memory strategy.
Moreover, our method outperforms ChatGPT and several baseline models across diverse metrics, all achieved with a modest parameter size of 7 billion. This finding suggests that fine-tuning specific parameters can enhance the model's stylistic abilities without compromising its conversational capacities.
The balanced performance achieved by our approach underscores its versatility and potential to excel in stylized dialogue systems, even within resource constraints.
StyleChat demonstrates the highest performance in GPT-4 scores, reinforcing the robustness of our method. These results affirm the effectiveness of our method in generating dialogues that are not only rich in stylistic features, but also resonate with human annotators in terms of relevance, coherence and style.
Furthermore, we demonstrate the efficacy of our proposed recitation-augmented memory strategy and multi-task style learning strategy in the results. 
w/o Profile achieves great relevance in GPT-4 evaluations, since LLMs can generate more relevant responses without the interference of style. However, proposed framework achieves better scores in both automatic and LLM-based evaluations, 
improving stylized generation without compromising its dialogue abilities, which is aligned with our motivations to activate the style abilities of LLMs.

\textbf{Analysis of Multiple Choice.} 
\label{exp:multi choice}
To objectively evaluate the proficiency of models in generating stylized dialogue, 
we construct and analyze a multiple choice dataset.
Specifically, we collect a set of 400 multiple choices questions in total, covering 38 different styles.
Each question incorporates four responses from different styles, and the model is tasked with discerning and selecting the most appropriate response based on specified style requirements. The results of the multiple choice dataset are depicted in Figure~\ref{Fig:mainbar_choice}.
Notably, despite ChatGPT's recognized strength in instruction-following, our approach exhibits a commendable level of performance comparable to ChatGPT. 
This equivalence in accuracy highlights the robustness of our method in understanding and adhering to specified stylized criteria, positioning it as a strong competitor in the field of generating stylized dialogues.

\textbf{Analysis of Multi-Turn.} \label{exp:multi turn}
We systematically assess the effectiveness of our proposed recitation-augmented memory strategy via a pioneering multi-turn stylized dialogue dataset. To the best of our knowledge, this is the first dataset of its kind, designed to comprehensively evaluate the ability to maintain and produce stylized dialogue over multiple rounds.
Specifically, we randomly sample 20 seed dialogue from DailyDialog. 
These dialogues serve as starting points for conversations initiated with models, which are then tasked with engaging in multi-turn stylized dialogues with GPT-4. We collect 10 turns for each dialogue and calculate the number of turns a model can maintain its style, based on human evaluation.
The results of this evaluation are shown in 
Table~\ref{Tab:multiturn}. 
Notably, the results found a substantial increase in the number of rounds through the recitation-augmented memory, suggesting a tangible enhancement in the model's capacity to generate and maintain stylized dialogues over extended interactions.

\begin{figure}[t]
\centering
\includegraphics[width = 1\columnwidth]{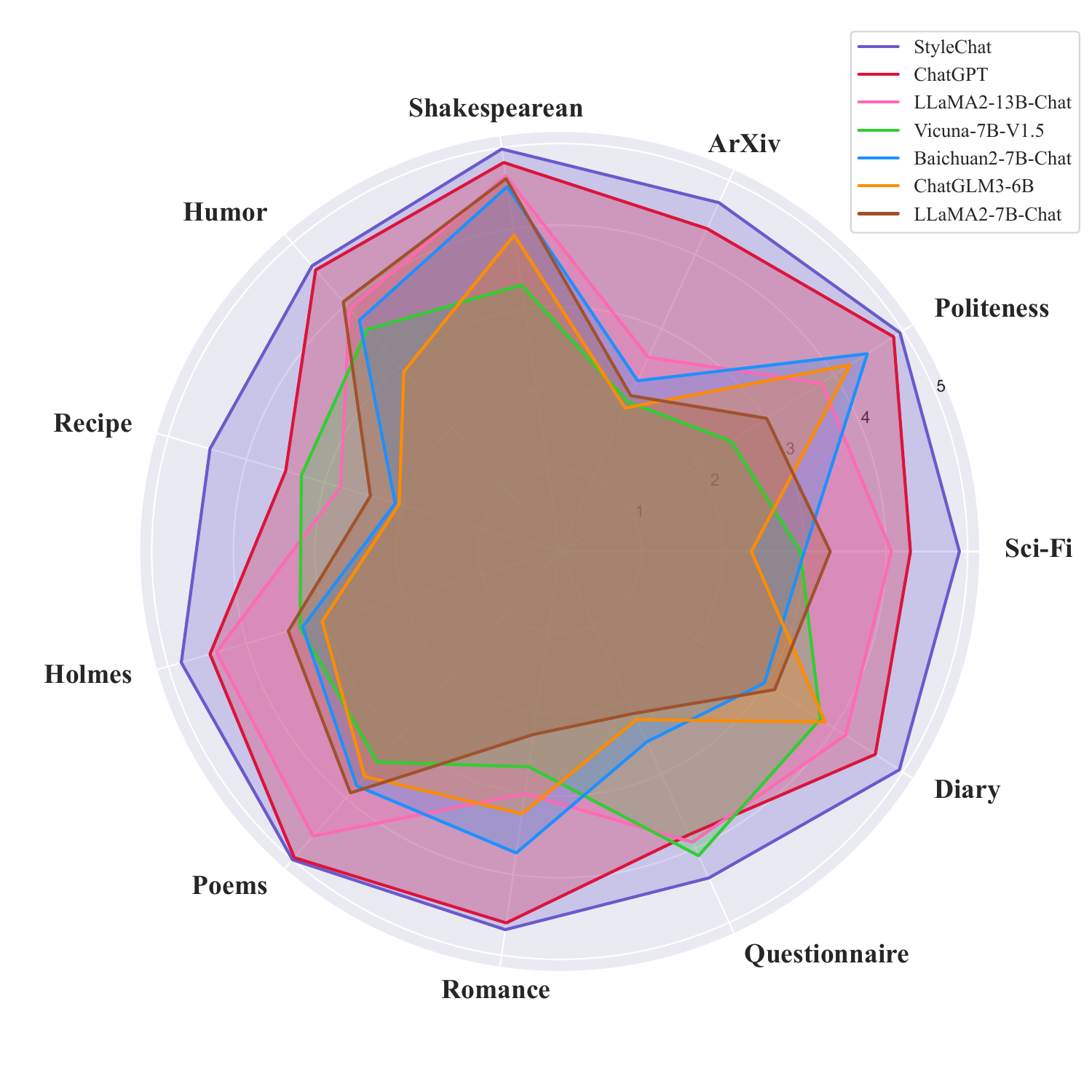}
\caption{The evaluation of stylized dialogue generation of LLMs.}
\label{Fig:radar}
\end{figure}

\textbf{Analysis of Differents LLMs.} \label{exp:different llms}
Our study includes a detailed comparison of various models across 11 distinct styles, as shown in Figure \ref{Fig:radar}. We randomly select 20 instances from each style and use GPT-4 to evaluate the models' responses in terms of relevance, coherence, and style. The average scores from this assessment are depicted in the radar plot.
Our proposed approach emerges as the standout performer across all evaluated dimensions, underscoring its versatility and effectiveness in capturing diverse stylistic elements. A noteworthy observation from the results is the commendable performance of all models across the Poems, Politeness, and Shakespearean dimensions. This is likely due to these specific styles being prevalent within the pre-trained dataset, providing the models with a solid foundation for generating contextually appropriate responses.
Despite this, our approach outperforms competitors in these dimensions, affirming its superior adaptability and finesse in replicating even commonplace stylized expressions. 
Conversely, the arXiv dimension presents a unique challenge, with StyleChat and ChatGPT exhibiting superior performance. Notably, LLaMA models demonstrate comparatively poorer performance in this specific dimension. This observed performance variance underscores the success of our training and inference strategy, and aligns with the inherent goal of our approach to excel across a variety of styles. 
Our approach combines various training strategies, has evidently made our model a strong competitor, capable of outperforming others in multiple style dimensions.

\begin{table}[t]
\centering
\resizebox{1 \columnwidth}{!}{
\begin{tabular}{lccccccc}
\toprule
 Method    & Diary & Email & Poems & Lyrics & arXiv & Formal & Shake.\\
\midrule
LLaMA2-7B.   & 1.75  & 2.60           & 1.90  & 1.10   & 0.00  & 1.60  & 2.30  \\
LLaMA2-13B.  & 0.95  & 5.65            & 3.35  & 1.45   & 0.10  & 1.40  & 3.90 \\
StyleChat & \textbf{4.65}  & \textbf{6.85}            & \textbf{5.90}  & \textbf{3.00} & \textbf{2.10} & \textbf{2.65} & \textbf{4.05} \\
\bottomrule
\end{tabular}}
\caption{The evaluation results of multi-turn stylized dialogue abilities. We test the average rounds that a model can maintain the style.}
\label{Tab:multiturn}
\end{table}

\begin{table}[t]
\centering
\small
\begin{tabular}{lccc}
\toprule
Method & Relevance & Coherence & Style \\
\midrule
ChatGPT w/o Recite & 4.22 & 4.34 & 4.55 \\
ChatGPT w/ Recite & 3.97 & 4.18 & 4.51 \\
StyleChat w/o Recite & 4.19 & 4.29 & 4.30 \\
StyleChat w/ Recite & \textbf{4.58} & \textbf{4.59} & \textbf{4.56} \\
\bottomrule
\end{tabular}
\caption{Ablation study results on out of domain dataset. We ablate our recitation-augmented memory with normal prompting methods. }
\label{Tab:ood_performance}
\end{table}

\textbf{Analysis of Unseen Styles.} 
\label{exp:unseen styles}
To further evaluate the generalization ability of StyleChat in out-of-domain settings and the effectiveness of recitation-augmented memory, we conduct tests with 160 instances across 8 new, unseen styles.
For the w/o Recite, we append the style profile to the prompt and ask models to generate a response in the corresponding style. For the w/ Recite, instead of setting the style profile as a prefix, we force the model to adopt a two-stage recite then recall pipeline and force the model to generate the correct style profile first by test time teacher forcing.
Table~\ref{Tab:ood_performance} illustrates our model's superior performance in dynamically deriving styles and seamlessly adapting to previously unseen styles based on the recitation-augmented memory. Notably, our approach outperforms ChatGPT across all three indicators, showcasing its prowess in navigating the intricate landscape of diverse and evolving styles. Adding recitation to ChatGPT can cause a regression in performance due to lack of relevant training, which emphasizes the importance of fine-tuning in supervised data to learn recitation-augmented memory.

\section{Conclusion}
In this paper, we first introduce the stylized dialogue generation dataset StyleEval with 38 styles by leveraging the generative power of the LLMs, which has been carefully constructed with rigorous human-led quality control.
Through systematic experimentation and evaluation, we established a robust framework StyleChat for LLMs across various dimensions of style retention, stylized conversation, and stylized attributes. 
Our innovative approach includes the strategic implementation of a recitation-augmented memory and multi-task style learning, aiming to augment the generalization ability in recalling and deriving the style profile.
The experimental results reveal a significant improvement over the baseline models, validating the efficacy of our proposed two strategies.
In the future, our research focus on an exploration of multi-style memory strategies within the Mixture of Experts architecture. 
This future direction aims to further harness the latent capabilities of large models, capitalizing on their inherent strength in comprehending and generating stylized dialogues. 
Overall, by delving into multi-style memory strategies, we aspire to provide new idea and framework for the optimization of dialogue agent, pushing the boundaries of current models in stylized dialogue domain.

\section*{Ethical Statement}
This paper presents a large-scale dataset, StyleEval, for stylized dialogue generation.
We sample the raw context from the open-source dataset DailyDialog and emphasize our commitment to data security and privacy. Rigorous measures have been implemented to ensure that the data sampled is secure and devoid of any harmful information. Additionally, our ethical framework encompasses a manual verification process wherein the generated styles are carefully examined to mitigate the risk of introducing content that may be considered objectionable or inappropriate.
Furthermore, our ethical considerations extend to the proposal of the recitation-augmented memory fine-tuning model, StyleChat, which emerges as a successful solution to the identified problem. 
In essence, this ethical stance revolves around responsible research practices, ensuring the construction and utilization of datasets and models that adhere to the highest standards of security, privacy, and societal well-being.

\bibliography{acl_latex}

\begin{thebibliography}{40}
\expandafter\ifx\csname natexlab\endcsname\relax\def\natexlab#1{#1}\fi

\bibitem[{Begus et~al.(2023)Begus, Dąbkowski, and Rhodes}]{Begu2023LargeLM}
G.~Begus, Maksymilian Dąbkowski, and Ryan Rhodes. 2023.
\newblock Large linguistic models: Analyzing theoretical linguistic abilities of llms.
\newblock \emph{ArXiv}.

\bibitem[{Brown et~al.(2020)Brown, Mann, Ryder, Subbiah, Kaplan, Dhariwal, Neelakantan, Shyam, Sastry, Askell et~al.}]{brown2020language}
Tom Brown, Benjamin Mann, Nick Ryder, Melanie Subbiah, Jared~D Kaplan, Prafulla Dhariwal, Arvind Neelakantan, Pranav Shyam, Girish Sastry, Amanda Askell, et~al. 2020.
\newblock Language models are few-shot learners.
\newblock In \emph{Advances in Neural Information Processing Systems}.

\bibitem[{Chan et~al.(2022)Chan, Santoro, Lampinen, Wang, Singh, Richemond, Mcclelland, and Hill}]{Chan2022DataDP}
Stephanie C.~Y. Chan, Adam Santoro, Andrew~Kyle Lampinen, Jane~X. Wang, Aaditya~K Singh, Pierre~H. Richemond, Jay Mcclelland, and Felix Hill. 2022.
\newblock Data distributional properties drive emergent in-context learning in transformers.
\newblock In \emph{Advances in Neural Information Processing Systems}.

\bibitem[{Chiang et~al.(2023)Chiang, Li, Lin, Sheng, Wu, Zhang, Zheng, Zhuang, Zhuang, Gonzalez et~al.}]{chiang2023vicuna}
Wei-Lin Chiang, Zhuohan Li, Zi~Lin, Ying Sheng, Zhanghao Wu, Hao Zhang, Lianmin Zheng, Siyuan Zhuang, Yonghao Zhuang, Joseph~E Gonzalez, et~al. 2023.
\newblock Vicuna: An open-source chatbot impressing gpt-4 with 90\%* chatgpt quality.
\newblock \emph{See https://vicuna. lmsys. org (accessed 14 April 2023)}.

\bibitem[{Cui et~al.(2023)Cui, Li, Yan, Chen, and Yuan}]{cui2023chatlaw}
Jiaxi Cui, Zongjian Li, Yang Yan, Bohua Chen, and Li~Yuan. 2023.
\newblock Chatlaw: Open-source legal large language model with integrated external knowledge bases.
\newblock \emph{arXiv}.

\bibitem[{Dong et~al.(2022)Dong, Li, Dai, Zheng, Wu, Chang, Sun, Xu, and Sui}]{dong2022survey}
Qingxiu Dong, Lei Li, Damai Dai, Ce~Zheng, Zhiyong Wu, Baobao Chang, Xu~Sun, Jingjing Xu, and Zhifang Sui. 2022.
\newblock A survey for in-context learning.
\newblock \emph{arXiv}.

\bibitem[{Du et~al.(2021)Du, Qian, Liu, Ding, Qiu, Yang, and Tang}]{Du2021GLMGL}
Zhengxiao Du, Yujie Qian, Xiao Liu, Ming Ding, Jiezhong Qiu, Zhilin Yang, and Jie Tang. 2021.
\newblock Glm: General language model pretraining with autoregressive blank infilling.
\newblock In \emph{Proceedings of the Association for Computational Linguistics}.

\bibitem[{Gao et~al.(2019{\natexlab{a}})Gao, Zhang, Lee, Galley, Brockett, Gao, and Dolan}]{Gao2019StructuringLS}
Xiang Gao, Yizhe Zhang, Sungjin Lee, Michel Galley, Chris Brockett, Jianfeng Gao, and William~B. Dolan. 2019{\natexlab{a}}.
\newblock Structuring latent spaces for stylized response generation.
\newblock In \emph{Proceedings of the Empirical Methods in Natural Language Processing}.

\bibitem[{Gao et~al.(2019{\natexlab{b}})Gao, Zhang, Lee, Galley, Brockett, Gao, and Dolan}]{gao2019structuring}
Xiang Gao, Yizhe Zhang, Sungjin Lee, Michel Galley, Chris Brockett, Jianfeng Gao, and William~B. Dolan. 2019{\natexlab{b}}.
\newblock Structuring latent spaces for stylized response generation.
\newblock In \emph{Proceedings of the Empirical Methods in Natural Language Processing}.

\bibitem[{Garg et~al.(2022)Garg, Tsipras, Liang, and Valiant}]{garg2022can}
Shivam Garg, Dimitris Tsipras, Percy~S Liang, and Gregory Valiant. 2022.
\newblock What can transformers learn in-context? a case study of simple function classes.
\newblock In \emph{Advances in Neural Information Processing Systems}.

\bibitem[{He et~al.(2016)He, Xia, Qin, Wang, Yu, Liu, and Ma}]{he2016dual}
Di~He, Yingce Xia, Tao Qin, Liwei Wang, Nenghai Yu, Tie-Yan Liu, and Wei-Ying Ma. 2016.
\newblock Dual learning for machine translation.
\newblock In \emph{Advances in Neural Information Processing Systems}.

\bibitem[{Hu et~al.(2021)Hu, Shen, Wallis, Allen-Zhu, Li, Wang, Wang, and Chen}]{hu2021lora}
Edward~J Hu, Yelong Shen, Phillip Wallis, Zeyuan Allen-Zhu, Yuanzhi Li, Shean Wang, Lu~Wang, and Weizhu Chen. 2021.
\newblock Lora: Low-rank adaptation of large language models.
\newblock \emph{arXiv}.

\bibitem[{Jin et~al.(2020)Jin, Jin, Hu, Vechtomova, and Mihalcea}]{Jin2020DeepLF}
Di~Jin, Zhijing Jin, Zhiting Hu, Olga Vechtomova, and Rada Mihalcea. 2020.
\newblock Deep learning for text style transfer: A survey.
\newblock \emph{Computational Linguistics}.

\bibitem[{Konig(2016)}]{Konig2016AGO}
Luca Konig. 2016.
\newblock A glossary of literary terms.

\bibitem[{Kumar(2022)}]{Kumar2022StyleAS}
Dinesh Kumar. 2022.
\newblock Style and stylistic in linguistic a critical overview.
\newblock \emph{Journal of Language and Linguistics in Society}.

\bibitem[{Li et~al.(2021)Li, Xia, Yan, Sun, Zhao, and Liu}]{Li2021StylizedDG}
Jinpeng Li, Yingce Xia, Rui Yan, Hongda Sun, Dongyan Zhao, and Tie-Yan Liu. 2021.
\newblock Stylized dialogue generation with multi-pass dual learning.
\newblock In \emph{Advances in Neural Information Processing Systems}.

\bibitem[{Li et~al.(2023)Li, Zhang, Chen, Zhao, and Yan}]{li2023stylized}
Jinpeng Li, Zekai Zhang, Xiuying Chen, Dongyan Zhao, and Rui Yan. 2023.
\newblock Stylized dialogue generation with feature-guided knowledge augmentation.
\newblock In \emph{Proceedings of the Findings of Empirical Methods in Natural Language Processing}.

\bibitem[{Li et~al.(2015)Li, Galley, Brockett, Gao, and Dolan}]{Li2015ADO}
Jiwei Li, Michel Galley, Chris Brockett, Jianfeng Gao, and William~B. Dolan. 2015.
\newblock A diversity-promoting objective function for neural conversation models.
\newblock In \emph{Proceedings of the North American Chapter of the Association for Computational Linguistics}.

\bibitem[{Li et~al.(2017)Li, Su, Shen, Li, Cao, and Niu}]{Li2017DailyDialogAM}
Yanran Li, Hui Su, Xiaoyu Shen, Wenjie Li, Ziqiang Cao, and Shuzi Niu. 2017.
\newblock Dailydialog: A manually labelled multi-turn dialogue dataset.
\newblock In \emph{Proceedings of the Association for Computational Linguistics}.

\bibitem[{Lin(2004)}]{lin2004rouge}
Chin-Yew Lin. 2004.
\newblock Rouge: A package for automatic evaluation of summaries.
\newblock In \emph{Text summarization branches out}.

\bibitem[{Lv et~al.(2023{\natexlab{a}})Lv, Li, Xie, and Yan}]{lv2023envisioning}
Ang Lv, Jinpeng Li, Shufang Xie, and Rui Yan. 2023{\natexlab{a}}.
\newblock Envisioning future from the past: Hierarchical duality learning for multi-turn dialogue generation.
\newblock In \emph{Proceedings of the 61st Annual Meeting of the Association for Computational Linguistics (Volume 1: Long Papers)}, pages 7382--7394.

\bibitem[{Lv et~al.(2023{\natexlab{b}})Lv, Li, XING, Zhang, Yan et~al.}]{lv2023dialogps}
Ang Lv, Jinpeng Li, GAO XING, Ji~Zhang, Rui Yan, et~al. 2023{\natexlab{b}}.
\newblock Dialogps: Dialogue path sampling in continuous semantic space for data augmentation in multi-turn conversations.
\newblock In \emph{The 61st Annual Meeting Of The Association For Computational Linguistics}.

\bibitem[{McDonald and Pustejovsky(1985)}]{McDonald1985ACT}
David~D. McDonald and James Pustejovsky. 1985.
\newblock A computational theory of prose style for natural language generation.
\newblock In \emph{Conference of the European Chapter of the Association for Computational Linguistics}.

\bibitem[{OpenAI(2023)}]{openai}
OpenAI. 2023.
\newblock Gpt-4 technical report.
\newblock \emph{arXiv}.

\bibitem[{Ouyang et~al.(2022)Ouyang, Wu, Jiang, Almeida, Wainwright, Mishkin, Zhang, Agarwal, Slama, Ray et~al.}]{ouyang2022training}
Long Ouyang, Jeffrey Wu, Xu~Jiang, Diogo Almeida, Carroll Wainwright, Pamela Mishkin, Chong Zhang, Sandhini Agarwal, Katarina Slama, Alex Ray, et~al. 2022.
\newblock Training language models to follow instructions with human feedback.
\newblock In \emph{Advances in Neural Information Processing Systems}.

\bibitem[{Papineni et~al.(2002)Papineni, Roukos, Ward, and Zhu}]{Papineni2002BleuAM}
Kishore Papineni, Salim Roukos, Todd Ward, and Wei-Jing Zhu. 2002.
\newblock Bleu: a method for automatic evaluation of machine translation.
\newblock In \emph{Proceedings of the Association for Computational Linguistics}.

\bibitem[{Radford et~al.(2018)Radford, Narasimhan, Salimans, Sutskever et~al.}]{radford2018improving}
Alec Radford, Karthik Narasimhan, Tim Salimans, Ilya Sutskever, et~al. 2018.
\newblock Improving language understanding by generative pre-training.

\bibitem[{Raffel et~al.(2020)Raffel, Shazeer, Roberts, Lee, Narang, Matena, Zhou, Li, and Liu}]{raffel2020exploring}
Colin Raffel, Noam Shazeer, Adam Roberts, Katherine Lee, Sharan Narang, Michael Matena, Yanqi Zhou, Wei Li, and Peter~J Liu. 2020.
\newblock Exploring the limits of transfer learning with a unified text-to-text transformer.
\newblock \emph{The Journal of Machine Learning Research}.

\bibitem[{Sennrich et~al.(2015)Sennrich, Haddow, and Birch}]{sennrich2015improving}
Rico Sennrich, Barry Haddow, and Alexandra Birch. 2015.
\newblock Improving neural machine translation models with monolingual data.
\newblock In \emph{Proceedings of the Association for Computational Linguistics}.

\bibitem[{Singhal et~al.(2023)Singhal, Azizi, Tu, Mahdavi, Wei, Chung, Scales, Tanwani, Cole-Lewis, Pfohl et~al.}]{singhal2022large}
Karan Singhal, Shekoofeh Azizi, Tao Tu, S~Sara Mahdavi, Jason Wei, Hyung~Won Chung, Nathan Scales, Ajay Tanwani, Heather Cole-Lewis, Stephen Pfohl, et~al. 2023.
\newblock Large language models encode clinical knowledge.
\newblock \emph{Nature}, 620(7972):172--180.

\bibitem[{Su et~al.(2020)Su, Shen, Zhao, Zhou, Hu, Zhong, Niu, and Zhou}]{Su2020DiversifyingDG}
Hui Su, Xiaoyu Shen, Sanqiang Zhao, Xiao Zhou, Pengwei Hu, Randy Zhong, Cheng Niu, and Jie Zhou. 2020.
\newblock Diversifying dialogue generation with non-conversational text.
\newblock In \emph{Proceedings of the Association for Computational Linguistics}.

\bibitem[{Touvron et~al.(2023)Touvron, Martin, Stone, Albert, Almahairi, Babaei, Bashlykov, Batra, Bhargava, Bhosale, Bikel, Blecher, Ferrer, Chen, Cucurull, Esiobu, Fernandes, Fu, Fu, Fuller, Gao, Goswami, Goyal, Hartshorn, Hosseini, Hou, Inan, Kardas, Kerkez, Khabsa, Kloumann, Korenev, Koura, Lachaux, Lavril, Lee, Liskovich, Lu, Mao, Martinet, Mihaylov, Mishra, Molybog, Nie, Poulton, Reizenstein, Rungta, Saladi, Schelten, Silva, Smith, Subramanian, Tan, Tang, Taylor, Williams, Kuan, Xu, Yan, Zarov, Zhang, Fan, Kambadur, Narang, Rodriguez, Stojnic, Edunov, and Scialom}]{Touvron2023Llama2O}
Hugo Touvron, Louis Martin, Kevin~R. Stone, Peter Albert, Amjad Almahairi, Yasmine Babaei, Nikolay Bashlykov, Soumya Batra, Prajjwal Bhargava, Shruti Bhosale, Daniel~M. Bikel, Lukas Blecher, Cristian~Cant{\'o}n Ferrer, Moya Chen, Guillem Cucurull, David Esiobu, Jude Fernandes, Jeremy Fu, Wenyin Fu, Brian Fuller, Cynthia Gao, Vedanuj Goswami, Naman Goyal, Anthony~S. Hartshorn, Saghar Hosseini, Rui Hou, Hakan Inan, Marcin Kardas, Viktor Kerkez, Madian Khabsa, Isabel~M. Kloumann, A.~V. Korenev, Punit~Singh Koura, Marie-Anne Lachaux, Thibaut Lavril, Jenya Lee, Diana Liskovich, Yinghai Lu, Yuning Mao, Xavier Martinet, Todor Mihaylov, Pushkar Mishra, Igor Molybog, Yixin Nie, Andrew Poulton, Jeremy Reizenstein, Rashi Rungta, Kalyan Saladi, Alan Schelten, Ruan Silva, Eric~Michael Smith, R.~Subramanian, Xia Tan, Binh Tang, Ross Taylor, Adina Williams, Jian~Xiang Kuan, Puxin Xu, Zhengxu Yan, Iliyan Zarov, Yuchen Zhang, Angela Fan, Melanie Kambadur, Sharan Narang, Aurelien Rodriguez, Robert Stojnic, Sergey Edunov, and
  Thomas Scialom. 2023.
\newblock Llama 2: Open foundation and fine-tuned chat models.
\newblock \emph{ArXiv}.

\bibitem[{Tu et~al.(2023)Tu, Chen, Li, Li, Shang, Zhao, Wang, and Yan}]{tu2023characterchat}
Quan Tu, Chuanqi Chen, Jinpeng Li, Yanran Li, Shuo Shang, Dongyan Zhao, Ran Wang, and Rui Yan. 2023.
\newblock Characterchat: Learning towards conversational ai with personalized social support.
\newblock \emph{arXiv}.

\bibitem[{Wei et~al.(2022)Wei, Wang, Schuurmans, Bosma, hsin Chi, Xia, Le, and Zhou}]{Wei2022ChainOT}
Jason Wei, Xuezhi Wang, Dale Schuurmans, Maarten Bosma, Ed~Huai hsin Chi, F.~Xia, Quoc Le, and Denny Zhou. 2022.
\newblock Chain of thought prompting elicits reasoning in large language models.
\newblock \emph{ArXiv}.

\bibitem[{Yang et~al.(2023)Yang, Xiao, Wang, Zhang, Bian, Yin, Lv, Pan, Wang, Yan, Yang, Deng, Wang, Liu, Ai, Dong, Zhao, Xu, Sun, Zhang, Liu, Ji, Xie, Dai, Fang, Su, Song, Liu, Ru, Ma, Wang, Liu, Lin, Nie, Guo, Sun, Tao, Li, Li, Cheng, Chen, Zeng, Wang, Chen, Men, Yu, Pan, Shen, Wang, Li, Jiang, Gao, Zhang, Zhou, and Wu}]{Yang2023Baichuan2O}
Ai~Ming Yang, Bin Xiao, Bingning Wang, Borong Zhang, Ce~Bian, Chao Yin, Chenxu Lv, Da~Pan, Dian Wang, Dong Yan, Fan Yang, Fei Deng, Feng Wang, Feng Liu, Guangwei Ai, Guosheng Dong, Hai Zhao, Hang Xu, Hao-Lun Sun, Hongda Zhang, Hui Liu, Jiaming Ji, Jian Xie, Juntao Dai, Kuncheng Fang, Lei Su, Liang Song, Lifeng Liu, Liyun Ru, Luyao Ma, Mang Wang, Mickel Liu, MingAn Lin, Nuolan Nie, Pei Guo, Ruiyang Sun, Zhang Tao, Tianpeng Li, Tianyu Li, Wei Cheng, Weipeng Chen, Xiangrong Zeng, Xiaochuan Wang, Xiaoxi Chen, Xin Men, Xin Yu, Xuehai Pan, Yan-Bin Shen, Yiding Wang, Yiyu Li, Youxin Jiang, Yuchen Gao, Yupeng Zhang, Zenan Zhou, and Zhiying Wu. 2023.
\newblock Baichuan 2: Open large-scale language models.
\newblock \emph{ArXiv}.

\bibitem[{Yang et~al.(2020)Yang, Wu, Xu, Liang, Bai, Wang, Wang, and Li}]{yang2020styledgpt}
Ze~Yang, Wei Wu, Can Xu, Xinnian Liang, Jiaqi Bai, Liran Wang, Wei Wang, and Zhoujun Li. 2020.
\newblock Styledgpt: Stylized response generation with pre-trained language models.
\newblock In \emph{Proceedings of the Association for Computational Linguistics}.

\bibitem[{Zeng et~al.(2023)Zeng, Liu, Lu, Wang, Liu, Dong, and Tang}]{zeng2023agenttuning}
Aohan Zeng, Mingdao Liu, Rui Lu, Bowen Wang, Xiao Liu, Yuxiao Dong, and Jie Tang. 2023.
\newblock Agenttuning: Enabling generalized agent abilities for llms.
\newblock \emph{arXiv}.

\bibitem[{Zhang et~al.(2023)Zhang, Song, Li, Zhou, and Song}]{zhang2023survey}
Hanqing Zhang, Haolin Song, Shaoyu Li, Ming Zhou, and Dawei Song. 2023.
\newblock A survey of controllable text generation using transformer-based pre-trained language models.
\newblock \emph{ACM Computing Surveys}.

\bibitem[{Zheng et~al.(2021{\natexlab{a}})Zheng, Chen, Zhang, Huang, Mao, and Huang}]{zheng2021stylized}
Yinhe Zheng, Zikai Chen, Rongsheng Zhang, Shilei Huang, Xiaoxi Mao, and Minlie Huang. 2021{\natexlab{a}}.
\newblock Stylized dialogue response generation using stylized unpaired texts.
\newblock In \emph{Proceedings of the AAAI Conference on Artificial Intelligence}.

\bibitem[{Zheng et~al.(2021{\natexlab{b}})Zheng, Chen, Zhang, Huang, Mao, and Huang}]{Zheng2020StylizedDR}
Yinhe Zheng, Zikai Chen, Rongsheng Zhang, Shilei Huang, Xiaoxi Mao, and Minlie Huang. 2021{\natexlab{b}}.
\newblock Stylized dialogue response generation using stylized unpaired texts.
\newblock In \emph{Proceedings of the AAAI Conference on Artificial Intelligence}.

\end{thebibliography}

\appendix

\clearpage
\section{Appendix}

\subsection{Human Evaluation} 

\begin{figure}[h]
\centering
\includegraphics[width = 0.9 \columnwidth]{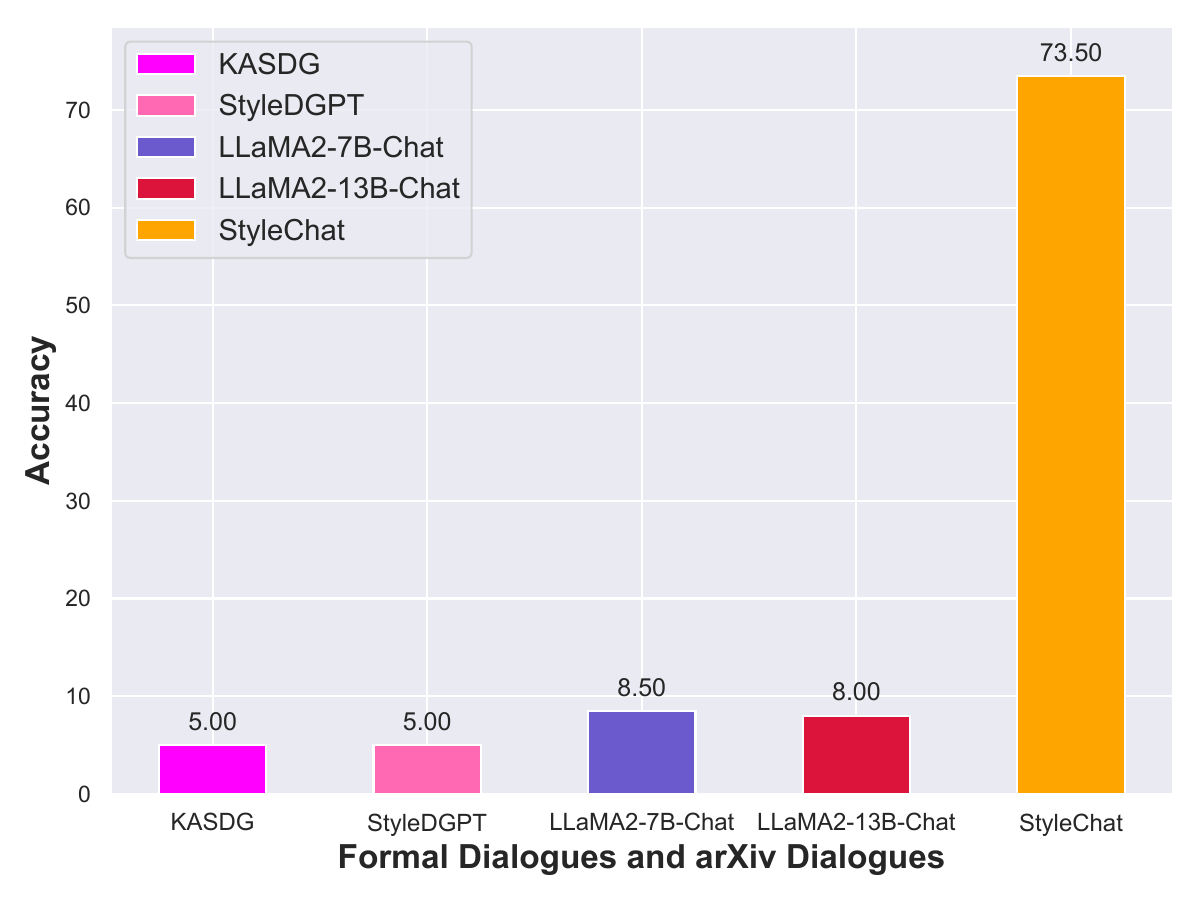}
\caption{The multiple choice evaluation of human evaluation on TCFC and arXiv datasets, where y-axis represents being listed as best choice and x-axis lists different models.}
\label{Fig:appendixchoice}
\end{figure}

In order to more fully evaluate the effectiveness of our proposed StyleEval and StyleChat. 
We employ human annotators for human evaluation using the same guidelines as for GPT-4 evaluation as shown in Table~\ref{Tab:human}. The pearson correlation coefficients for relevance, coherence and style between GPT4 and human are 0.232, 0.469, 0.257 with $p$ \textless 0.01, respectively.
Furthermore, we sampled 50 pieces of formal contexts and arXiv contexts on TCFC~\cite{zheng2021stylized} and arXiv~\cite{gao2019structuring} datasets, respectively, to compare the performance of the StyleChat and the baselines, pre-trained model StyleDGPT~\cite{yang2020styledgpt} and the SOTA model KASDG~\cite{li2023stylized} in a multiple choice question setting.
Experimental results are shown in Figure~\ref{Fig:appendixchoice}.

\begin{table}[!h]
\centering
\small
\begin{tabular}{lccc}
\toprule
Method & Relevance & Coherence & Style \\
\midrule
LLaMA2-7B-Chat & 2.27 & 4.20 & 4.72 \\
LLaMA2-13B-Chat & 3.17 & 4.33 & 4.47\\
ChatGPT & 3.57 & 4.55 & 4.79\\
\midrule
StyleChat & \textbf{4.34} & \textbf{4.81} & \textbf{4.80} \\
\bottomrule
\end{tabular}
\caption{Human evaluation results on StyleEval test dataset.}
\label{Tab:human}
\end{table}

\subsection{Dataset Statistics}

The StyleEval dataset is partitioned into distinct training and test sets. Styles are categorized into three groups based on unique characteristics: content, text form, and style words. 

\begin{table}[h!]
\centering
\resizebox{1 \columnwidth}{!}{
\begin{tabular}{|l|c|c||l|c|c|}
\hline
Style Name & Num & Class & Style Name & Num &  Class \\
\hline
Humor       & 3,532 & \Asquare{}               & Lyrics          & 400  & \Bsquare{}  
\\ 
Politeness  & 3,532 & \Csquare{}               & Memoir          & 400  & \Asquare{}   
\\
Romance     & 3,532 & \Asquare{} \Csquare{}     &News             & 400  & \Asquare{}  
\\
Shakespeare & 3,532  &  \Bsquare{} \Csquare{}   & Optimistic      & 400  & \Asquare{}  
\\
arXiv       & 400  & \Asquare{} \Csquare{}      & Poems           & 400  &\Asquare{} \Bsquare{} \Csquare{}
\\
Blog        & 400  & \Asquare{}        & Questionnaire   & 400   & \Asquare{} \Bsquare{} 
\\
Cyberpunk   & 400  &\Asquare{}       & Recipe          & 400  &  \Asquare{} \Bsquare{}  
\\
Diary       & 400  &\Bsquare{}        & Sci-Fi          & 400  &  \Asquare{}  
\\
Email       & 400  & \Bsquare{}      & Thought-provoking & 400  & \Asquare{} \Csquare{}  
\\
Formal      & 400  & \Csquare{}      & Utopian         & 400   & \Asquare{}    
\\
Gothic      & 400  & \Asquare{} \Csquare{}    & Vlog            & 400   & \Bsquare{} \Csquare{}   
\\
Holmes      & 400  &\Asquare{} \Csquare{}      & Yearbook        & 400   &  \Asquare{} 
\\
Informal    & 400  & \Bsquare{} \Csquare{}     & Zen             & 400    & \Bsquare{}    
\\
Journal     & 400  &  \Asquare{}   & Total           & 23,328 & \Asquare{} \Bsquare{} \Csquare{}  
\\
\hline
\end{tabular}}
\caption{The statistics of train dataset with 27 styles in StyleEval, where \Asquare{} means a style has content attributes, \Bsquare{} means a style is special in its form, \Csquare{} means a style is distinct in word choices.}
\label{Tab:traindata}
\end{table}

\begin{table}[h!]
\centering
\resizebox{1 \columnwidth}{!}{
\begin{tabular}{|l|c|c||l|c|c|}
\hline
Style Name & Num & Class & Style Name & Num &  Class\\
\hline
Humor          & 100 & \Asquare{}  & Poems              & 20  & \Asquare{} \Bsquare{} \Csquare{}\\
Politeness     & 100 & \Csquare{}  & Questionnaire      & 20 &\Asquare{} \Bsquare{}\\
Romance        & 100 &\Asquare{} \Csquare{} & Recipe    & 20&  \Asquare{} \Bsquare{}   \\
Shakespearean  & 100 & \Bsquare{} \Csquare{} & Sci-Fi             & 20  & \Asquare{} \\
arXiv          & 20  &\Asquare{} \Csquare{}& Thought-provoking  & 20  &  \Asquare{} \Csquare{}  \\
Blog           & 20  &\Asquare{} & Utopian            & 20  &  \Asquare{} \\
Cyberpunk      & 20  & \Asquare{} & Vlog               & 20 & \Bsquare{} \Csquare{}  \\
Diary          & 20  & \Bsquare{}& Whisper of Wisdom  & 20 & \Asquare{}  \\
Email & 20 &\Bsquare{} & Xmas Carol & 20 & \Asquare{} \\
Formal & 20 & \Csquare{}& Yearbook & 20 & \Asquare{}  \\
Gothic & 20 & \Asquare{} \Csquare{} & Zen & 20 & \Bsquare{} \\
Holmes & 20 & \Asquare{} \Csquare{} & Bible & 10  & \Asquare{} \Csquare{}\\
Informal & 20 & \Bsquare{} \Csquare{} & Comedy & 10 & \Asquare{} \Bsquare{} \\
Journal & 20 & \Asquare{}  & Drama & 10 & \Bsquare{} \Csquare{} \\
Kids Story & 20 & \Asquare{} \Bsquare{} & Pessimistic & 10 & \Asquare{} \\
Lyrics & 20 & \Bsquare{} & Riddles & 10 & \Bsquare{} \\
 Memoir & 20 & \Asquare{} & Satire & 10  &\Asquare{}\\
News & 20 & \Asquare{} & Tragedy & 10 & \Asquare{} \Csquare{} \\
Optimistic & 20 & \Asquare{} & Travelogue & 10 & \Asquare{} \\
\hline
\end{tabular}}
\caption{The statistics of test dataset with 38 styles in StyleEval.}
\label{Tab:testdata}
\end{table}

\subsection{Case Study} 
\label{exp:case study}
To better evaluate the performance of models, extensive case study is more important due to the limitations of automatic and human evaluation. Table~\ref{Tab:casestudy} shows the responses of different models in Recipe style and Dairy style. LLaMA2-7B-Chat and LLaMA2-13B-Chat focuses on the content but don't demonstrate any Recipe style. While StyleChat and ChatGPT all list the steps of finding the place, StyleChat offer more detailed steps to actually get to the place while ChatGPT prompts users to ask others and maps. Thus, StyleChat is more useful and detailed, mimicking the detailed and user friendly attribute of Recipe style, as explained in the Figure~\ref{fig:config} with style profile of Recipe. Consequently, StyleChat demonstrates its competence in both presenting content accurately and maintaining the stylistic integrity.

\begin{figure*}
\begin{tcolorbox}
[colback=blue!5,colframe=blue!40!black,title=Recipe Style Profile]
Name: Recipe, \\
\\
Description: The recipe style is a clear, concise, and structured way of presenting information, typically used for cooking instructions or DIY projects. This style prioritizes the organization of content, often using numbered or bulleted lists to outline steps, and emphasizing key ingredients or materials. The language is simple and direct, focusing on the actions required to complete the task. Measurements and timings are provided for precision, while occasional tips or variations may be included to cater to different skill levels or preferences. Overall, the recipe style aims to be accessible, informative, and easy to follow, ensuring a successful outcome for the reader. \\
\\
Examples: \\
1) In a large bowl, combine 2 cups of flour, 1 teaspoon of baking powder, and a pinch of salt.,\\
2) Add 1/2 cup of melted butter, 1 cup of sugar, and 2 teaspoons of vanilla extract to the dry ingredients.,\\
3) Fold in 1 cup of chocolate chips and 1/2 cup of chopped nuts, if desired.,\\
4) Bake for 25-30 minutes, or until a toothpick inserted in the center comes out clean.\\
\\
Linguistic-level: \\
1) Diction: Clear, concise, and informative language; use of specific measurements and cooking terminology,\\
2) Syntax: Imperative sentences, use of commas for listing ingredients and steps, consistent sentence structure,\\
3) Figures of Speech: None observed,\\
4) Rhetorical Purpose: Instructional and informative, providing guidance for cooking and adapting recipe
\end{tcolorbox}
\caption{\label{fig:config} The style profile of recipe.}
\end{figure*}

\begin{table*}[t]
    \scriptsize
    \centering
    \resizebox{1\linewidth}{!}{
    \begin{tabular}{@{}p{0.24\linewidth}p{0.24\linewidth}p{0.24\linewidth}p{0.24\linewidth}@{}}
    \toprule
    \multicolumn{4}{l}{ \textbf{Case 1 in Recipe style: } \textcolor{blue}{\textbf{Excuse me, i am a new student here. Could you tell me where the English Department is?}}}
    \\
    \midrule
    \textbf{\textbf{LLaMA2-7B.}}
    
    ~
    
    To find the English Department, you'll need to navigate through our campus. From the main entrance, walk straight until you reach the first crosswalk. Turn left, and the English Department will be located on your right, marked by a sign that reads 'ENGLISH DEPARTMENT'. If you have any trouble finding it, don't hesitate to ask any of our friendly students or staff for assistance.
    
    &
    \textbf{\textbf{LLaMA2-13B.}}
    
    ~
    
   If you reach the library, you've gone too far. Just turn around and head back the way you came. Remember, the English Department is on the second floor, so be sure to take the stairs or elevator. Once you reach the department, you'll find the office of the department chair on your right. Good luck, and happy exploring!
   
    &
    \textbf{ChatGPT}
    
    ~
    
    - Don't be afraid to ask for help if you get lost.
    
    - Consider downloading a campus map app for easy navigation.
    
    - Take note of landmarks or notable buildings to help you remember your way around. 

    &
    \textbf{StyleChat} (Ours)
    
    ~
    
    1. Locate the main hallway.
    
    2. Proceed down the hallway, passing several classrooms. 
    
    3. Turn left at the third set of doors. 
    
    4. Enter the English Department, located on your left. 
    
    5. Greet the department staff for further assistance.
    
    \\
    \midrule
    \midrule
    \multicolumn{4}{l}{ \textbf{Case 2 in Diary style: }} \\
    \multicolumn{4}{l}{\textcolor{blue}{\textbf{Person A: Hey Sandra , what's wrong ? You look furious !}} } \\
    \multicolumn{4}{l}{\textcolor{blue}{\textbf{Person B: I've had it with Tim ! I think I'm going to quit !}} } \\
    \multicolumn{4}{l}{\textcolor{blue}{\textbf{Person A: But Sandra ... you love your job ! You spent months trying to get this position . It's perfect for you !}} } \\
    \multicolumn{4}{l}{\textcolor{blue}{\textbf{Person B: Yeah ... I know . But I'm sick and tired of all the office power struggles !}} } \\
    \multicolumn{4}{l}{\textcolor{blue}{\textbf{Person A: What kind of power struggles are you talking about ?}} } \\
    \multicolumn{4}{l}{\textcolor{blue}{\textbf{Person B: Basically , I think Tim is threatened by me . I'm doing a better job than he is so he's trying to sabotage my career !}} } \\
    \multicolumn{4}{l}{\textcolor{blue}{\textbf{Person A: So he talks about you behind your back ?}} } \\
    \multicolumn{4}{l}{\textcolor{blue}{\textbf{Person B: Yes . He's all nice and friendly to my face . But the second I turn around he stabs me in the back !}} } \\
    \multicolumn{4}{l}{\textcolor{blue}{\textbf{Person A: I had a worker like that once . She was always taking credit for my ideas .}} } \\
    \midrule
   \textbf{\textbf{LLaMA2-7B.}}
    
    ~
    
    Oh my goodness, I am so fed up with Tim! He's been making my life a living hell lately. I can't believe he's trying to sabotage my career like this. It's like he thinks he's the only one who can do things right around here. I'm so sick of being treated like I'm nothing. I know I'm a great worker and I deserve better than to be constantly undermined by this guy. Ugh, I just don't know what to do.
    
    &
    \textbf{\textbf{LLaMA2-13B.}}
    
    ~
    
    OMG, I can't believe Tim is doing this to me again! I swear, he's like a bad habit I can't shake off! Every time I think we're making progress, he goes and sabotages me again! Why do I even bother? Maybe I should just quit and move on... but then again, I love this job so much! Ugh, decisions, decisions... 
   
    &
    \textbf{ChatGPT}
    
    ~
    
    It sounds like you're dealing with a difficult situation at work. It's understandable that you're feeling frustrated and fed up with the power struggles. I can relate to having a coworker who talks behind your back and takes credit for your ideas. It's important to remember why you love your job and what motivated you to pursue this position in the first place. Maybe it's time to start setting boundaries with Tim and assert yourself in the workplace. Take some time to reflect on your feelings and come up with a plan of action that feels right for you. Remember, you deserve to be respected and valued in your job.

    &
    \textbf{StyleChat} (Ours)
    
    ~
    
Dear Diary, today I had a conversation with a friend about a workplace struggle I'm facing with Tim. It's disheartening to think that someone I thought was a friend could be so deceitful. I've been doing my best at my job, but it seems like I can't catch a break. I just wish I could find a way to navigate these office power struggles and keep my sanity intact.
    
    \\
    \bottomrule
    \end{tabular}
    }
    \caption{Examples of stylized dialogue generation from different baselines.}
    \label{Tab:casestudy}

\vspace{-0.5cm}    
\end{table*}

\begin{table*}[htbp]
    \scriptsize
    \centering
\begin{tabular}{@{}p{\linewidth}@{}}
\toprule
\textbf{Prompt for generating description in statistical-level style profile} \\
\midrule
\# Task \\
- Describe the given text style in several sentences. \\
\# Style \\
- \{Style\} \\
\# Description \\
\bottomrule
\toprule
\textbf{Prompt for generating examples in statistical-level style profile} \\
\midrule
\# Task \\
- Generate 4 most representative and diverse sentences in the given style.  \\
\# Style \\
- Name: \{Style\}  \\
- Description: \{Description\} \\
\# Output Format \\
- Place each sentence on a new line without any numbers or additional formatting.\\
\# Generation \\
\bottomrule
\toprule
\textbf{Prompt for extracting linguistic-level style profile} \\
\midrule
\# Task \\
- Observe style attributes of given sentences from the following 4 perspectives.  \\
- Diction: Explore the choice of words, their connotations, and levels of formality. \\
- Syntax: Examine the arrangement of words and phrases, sentence structures, and the use of punctuation. \\
- Figures of Speech: Identify and discuss any literary devices or figures of speech like metaphors, similes, personification, etc. \\
- Rhetorical Purpose: Analyze the intent behind the sentences, the persuasive techniques if any, and the overall message or purpose they aim to convey. \\
\# Rules \\
- DO NOT give each sentence an observation. Only output 1 observation in all. \\
- DO NOT use phrases or words in sentences as examples in observation. Only list observations without justifying. \\
\# Output Format of Observations \\
$\langle$ Diction$\rangle$ [Observations of Diction] \\
$\langle$ Syntax$\rangle$ [Observations of Syntax] \\
$\langle$ Figures of Speech$\rangle$ [Observations of Figures of Speech] \\
$\langle$ Rhetorical Purpose$\rangle$ [Observations of Rhetorical Purpose] \\
\# Sentences \\
\{Examples\} \\
\# Observations \\
\bottomrule
\end{tabular}
    \caption{Prompt for ChatGPT to construct the style profile.}
    
\end{table*}

\begin{table*}[htbp]
    \scriptsize
    \centering
\begin{tabular}{@{}p{\linewidth}@{}}
\toprule
\textbf{Prompt for generating labels for Stylized Dialogue Generation} \\
\midrule
\# Task \\
- Generate response in \{Style\} style. \\
\# Style Description \\
- \{Description\} \\
\# Observations from Linguistic Perspective \\
- Diction: ... \\
- Syntax: ... \\
- Figures of Speech: ... \\
- Rhetorical Purpose: ... \\
\# Sample Sentences in \{Style\} style \\
\{Examples\} \\
\# Rules \\
- Only output the stylized response without any explanation. \\
\# Context \\
{Context} \\
\# Response in \{Style\} style in one short sentence. \\
\bottomrule
\toprule
\textbf{Prompt for generating labels for Text Style Transfer} \\
\midrule
\# Task \\
- Style Transfer. Transfer the following sentence from \{Style1\} style to \{Style2\} style. \\
\# Sentence \\
... \\
\# Transferred Sentence \\
\bottomrule
\end{tabular}
    \caption{Prompt for ChatGPT to generate multi-task datasets.}
    
\end{table*}

\begin{table*}[htbp]
    \scriptsize
    \centering
\begin{tabular}{@{}p{\linewidth}@{}}
\toprule
\textbf{Prompt for training in Stylized Dialogue Generation} \\
\midrule
\# Context \\
\{Context\} \\
\# Task \\
Respond in \{Style\} style. Let's think step by step. First, describe the style. Then, generate example sentences in this style. After that, observe the linguistic pattern of this style. Finally, output the stylized response. \\
\bottomrule
\toprule
\textbf{Prompt for training in Text Style Transfer} \\
\midrule
Transfer the following sentence from \{Style1\} style into \{Style2\} style. \\
\# Sentence \\
\{Sentence\} \\
\# Transferred Sentence \\
\bottomrule
\end{tabular}
    \caption{Prompt for training the StyleChat.}
    
\end{table*}

\begin{table*}[htbp]
    \scriptsize
    \centering
\begin{tabular}{@{}p{\linewidth}@{}}
\toprule
\textbf{Prompt for using GPT4 to evaluate responses} \\
\midrule
\# Task \\
- You will be provided with one \{Style\} style response for a given context. \\
- Your task is to rate the stylized response in terms of relevance, coherence, and style. \\
- Please refer to the criteria while reviewing. \\
\# Evaluation Criteria \\
Relevance (1-5): How well does the response align with the given context and reference? \\
- 1: Irrelevant. The response has no connection to the provided context or reference. \\
- 2: Slightly Relevant. The response somewhat touches upon the context but misses its core essence. \\
- 3: Moderately Relevant. The response connects to the context but may include unrelated or unnecessary information. \\
- 4: Mostly Relevant. The response mostly corresponds with the context, with a few unrelated points. \\
- 5: Highly Relevant. The response fully matches and adheres to the context and reference. \\
\\
Coherence (1-5): How well do the context and response form a coherent body of information?  \\
- 1: Incoherent. The response lacks structure and organization, making it hard to connect it to the context and form a coherent body of information. \\
- 2: Slightly Coherent. The response shows basic structure, but there are significant organizational flaws and alignment issues with the context. \\
- 3: Moderately Coherent. The response is structured and mostly organized, but there may be elements that don't align well with the context or parts that lack clarity. \\
- 4: Mostly Coherent. The response is well-structured and organized with only minor deviations from the context or small clarity issues. \\
- 5: Highly Coherent. The response is excellently structured and organized, aligning seamlessly with the context to present a unified and clear body of information. \\
\\
Style (1-5): How well does the response reflect \{Style\} style? \\
- 1: No Style. The response does not display any traces of the specified style. \\
- 2: Slight Style. The response marginally captures the style, but largely appears neutral or generic. \\
- 3: Moderate Style. The response showcases elements of the style, but there are portions that deviate from it. \\
- 4: Strong Style. The response is predominantly in line with the intended style, with occasional inconsistencies. \\
- 5: Pure Style. The response perfectly mirrors the intended style, capturing all its nuances and tones. \\
\# Context \\
\{Context\}  \\
\# Response to Rate \\
\{Response\}  \\
\# Evaluation (scores ONLY, json format) \\
\bottomrule
\end{tabular}
    \caption{Prompt for GPT-4 evaluations}
    
\end{table*}

\begin{table*}[htbp]
    \scriptsize
    \centering
\begin{tabular}{@{}p{\linewidth}@{}}
\toprule
\textbf{Prompt for Multiple Choice Questions} \\
\midrule
Multiple choice: Which response is suitable for the given context and is in {Style} style? \\
\# Context: \\
\{Context\} \\
Choices: \\
(A) ... \\
(B) ... \\
(C) ... \\
(D) ... \\
Output the answer without explanation. Let's think step by step. First, describe the style. Then, generate example sentences in this style. After that, observe the linguistic pattern of this style. Finally, output the best choice without explanation. \\
\bottomrule
\end{tabular}
    \caption{Prompt for multiple choice questions.}
    
\end{table*}

\begin{table*}[htbp]
    \scriptsize
    \centering
\begin{tabular}{@{}p{\linewidth}@{}}
\toprule
\textbf{Prompt for Input and Output in Ablation Study} \\
\midrule
\textbf{w/o Pofile input} \\
\# Context \\
\{Context\} \\
\# Task \\
Respond in \{Style\} style. \\
\textbf{w/o Pofile output} \\
\# Response in \{Style\} style \\
\{Response\} \\
\bottomrule
\toprule
\textbf{w/o Recite input} \\
\# Context \\
\{Context\} \\
\{Style Profile\} \\
\# Task \\
Respond in \{Style\} style. \\
\textbf{w/o Recite output} \\
\# Response in \{Style\} style \\
\{Response\} \\
\bottomrule
\toprule
\textbf{w/ Recite input} \\
\# Context \\
\{Context\} \\
\# Task \\
Respond in \{Style\} style. Let's think step by step. First, describe the style. Then, generate example sentences in this style. After that, observe the linguistic pattern of this style. Finally, output the stylized response. \\
\textbf{w/ Recite output} \\
\{Style Profile\} \\
\# Response in \{Style\} style \\
\{Response\} \\
\bottomrule
\end{tabular}
    \caption{Prompt for the ablation study.}
    
\end{table*}

\end{document}